\documentclass[10pt,twocolumn,letterpaper]{article}

\usepackage[pagenumbers]{cvpr} 

\usepackage{graphicx}
\usepackage{amsmath}
\usepackage{amssymb}
\usepackage{booktabs}

\usepackage{times}
\usepackage{epsfig}
\usepackage{array}
\usepackage{overpic} 
\newlength\savewidth

\usepackage[pagebackref,breaklinks,colorlinks]{hyperref}

\usepackage[capitalize]{cleveref}
\crefname{section}{Sec.}{Secs.}
\Crefname{section}{Section}{Sections}
\Crefname{table}{Table}{Tables}
\crefname{table}{Tab.}{Tabs.}

\def\cvprPaperID{6141} 
\def\confName{CVPR}
\def\confYear{2022}

\begin{document}

\title{Exemplar-free Class Incremental Learning via \\
Discriminative and Comparable One-class Classifiers}

\author{Wenju Sun \ \ \ \ Qingyong Li \ \ \ \ Jing Zhang \ \ \ \ Danyu Wang \ \ \ \ Wen Wang \ \ \ \ Yangli-ao Geng\thanks {Corresponding author: Yangli-ao Geng (gengyla@bjtu.edu.cn).}\\
Beijing Key Lab of Traffic Data Analysis and Mining, Beijing Jiaotong University, Beijing, 100044, China\\
{\tt\small \{SunWenJu,liqy,j\_zhang,WangDanYu,wangwen,gengyla\}@bjtu.edu.cn}
}

\maketitle

\begin{abstract}
  The exemplar-free class incremental learning requires classification models to learn new class knowledge incrementally without retaining any old samples. Recently, the framework based on parallel one-class classifiers (POC), which trains a one-class classifier (OCC) independently for each category, has attracted extensive attention, since it can naturally avoid catastrophic forgetting. POC, however, suffers from weak discriminability and comparability due to its independent training strategy for different OOCs. To meet this challenge, we propose a new framework, named \textbf{Dis}criminative and \textbf{C}omparable \textbf{O}ne-class classifiers for \textbf{I}ncremental \textbf{L}earning (DisCOIL). DisCOIL follows the basic principle of POC, but it adopts variational auto-encoders (VAE) instead of other well-established one-class classifiers (e.g. deep SVDD), because a trained VAE can not only identify the probability of an input sample belonging to a class but also generate pseudo samples of the class to assist in learning new tasks. With this advantage, DisCOIL trains a new-class VAE in contrast with the old-class VAEs, which forces the new-class VAE to reconstruct better for new-class samples but worse for the old-class pseudo samples, thus enhancing the comparability. Furthermore, DisCOIL introduces a hinge reconstruction loss to ensure the discriminability. We evaluate our method extensively on MNIST, CIFAR10, and Tiny-ImageNet. The experimental results show that DisCOIL achieves state-of-the-art performance. The source code will be publicly available. \footnote {***/***/***}
\end{abstract}

\section{Introduction\label{section1}}
In an ever-changing environment, we would like an intelligent system to learn new knowledge incrementally without forgetting, which refers to incremental learning (IL) \cite{survey1, survey2}.
IL aims to learn knowledge from a sequence of disjoint tasks.
According to whether the task identification of samples is available during the test stage, there exist two IL settings: Task-IL (task identification available) and Class-IL (task identification unavailable) \cite{setting}. 
As task identifications are usually unavailable in practice, Class-IL has received more attention in recent years. 
Furthermore, it is often desired that the samples of old tasks are not stored for the consideration of data security and privacy \cite{pcl}, which motivates research in exemplar-free Class-IL \cite{classilsurvey}. Our paper focuses on this IL setting.

\begin{figure}[t]
    \centering
    \includegraphics[width=1.0\columnwidth]{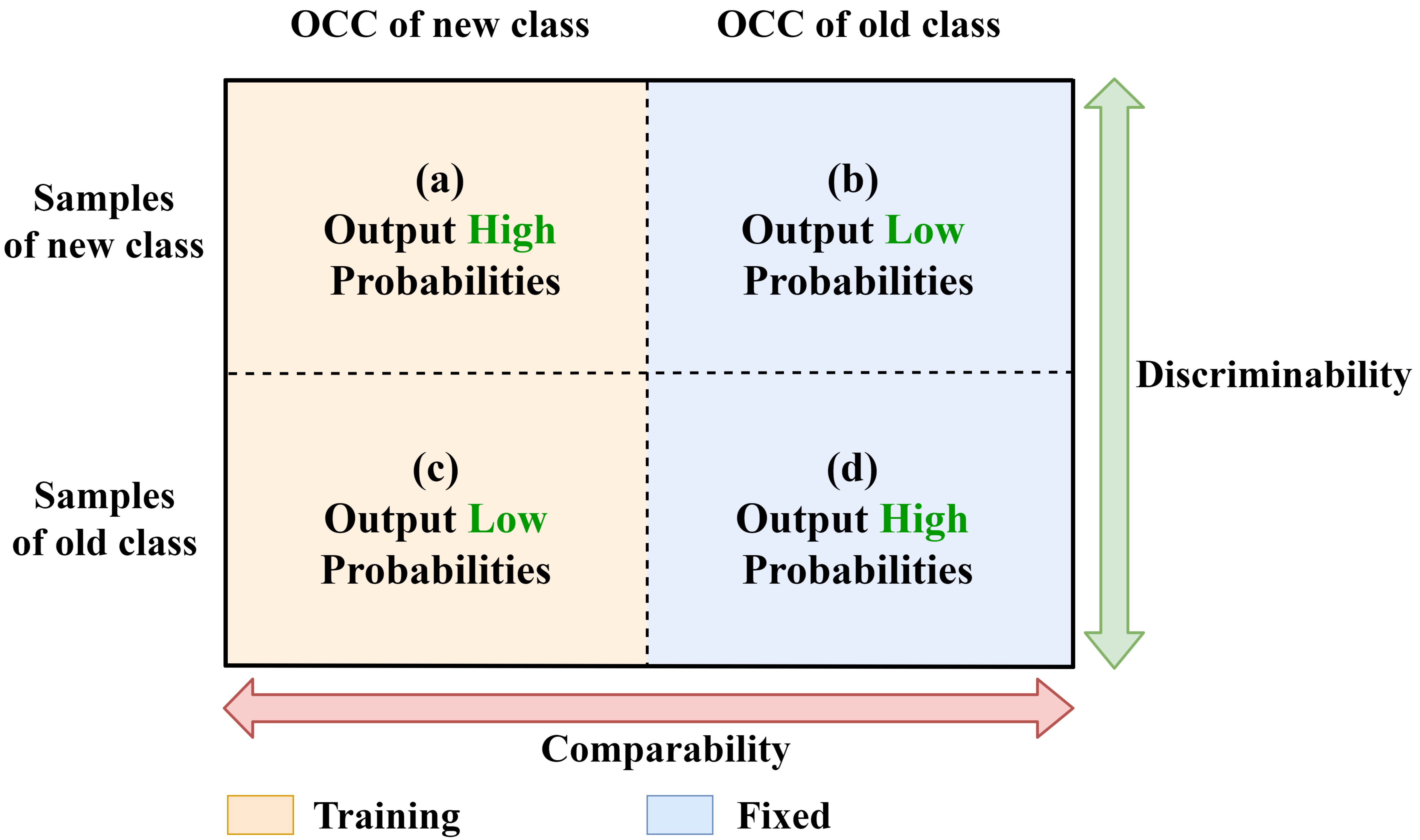} 
    \caption{
        Ideal incremental learning results with two one-class classifiers (OCC).
        Following the vertical arrow viewpoint, both OCCs should have discriminability to determine whether an input sample belongs to their corresponding class.
        Following the horizontal arrow viewpoint, they need to have comparability that allows the outputs of the two OCCs to be compared to each other.
    }
    \label{fig:comparability}
\end{figure}

Exemplar-free class-IL requires the model to learn a new class knowledge without any sample information of old classes, which is quite challenging and leads many established IL algorithms to fail under this scenario \cite{der, pcl, ilcoc}. One reason is that the commonly used softmax classifier is prone to overfit the new-task samples without referring to old-task samples, making the classification output bias towards the new-task classes even for the old-task input, which is known as the forgetting problem \cite{forget1, forget2}. To tackle this issue, several algorithms following the framework of parallel one-class classifiers (POC) have been proposed \cite{ilcoc, pcl}. They train a new one-class classifier (OCC) for each class and thus disentangle the interference of the new task to the old tasks. With this design, the POC-based algorithms naturally avoid the forgetting problem. They, however, are vulnerable to poor performance due to weak discriminability within each OCC and comparability among different OCCs.

\begin{figure}[t]
  \centering
  \includegraphics[width=0.9\columnwidth]{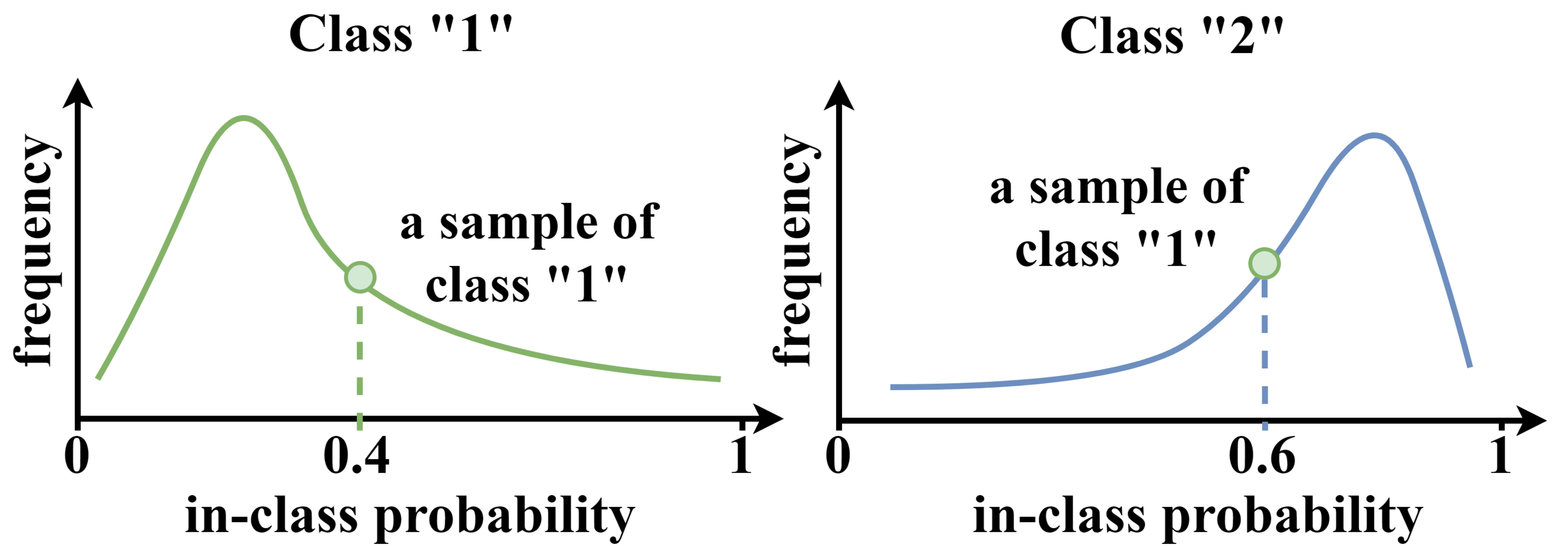} 
  \caption{
    An example of misclassification that caused by two non-comparable one-class classifiers. The two curves represent the output distribution of the two discriminative one-class classifiers.
  }
  \label{fig:distribution}
\end{figure}

More specifically, we illustrate how the discriminability and the comparability affect the performance of POC-based methods in Figure~\ref{fig:comparability}. Without loss of generality, let us consider a POC architecture consisting of two OCCs, where one is the current training OCC for a new class and the other is the fixed OCC for an old class. From the vertical viewpoint (green arrow), both the new-class OCC and the old-class OCC should output high probabilities for their corresponding class samples while low probabilities for the others, which we call the \textbf{discriminability} within each OCC. On the other hand, from the horizontal viewpoint (red arrow), for new-class samples, the new-class OCC should output higher probabilities than the old-class OCC; for old-class samples, the probabilities output by the new-class OCC should be lower than that of the old-class OCC. We refer to this property as the \textbf{comparability} among OCCs. 
The POC framework usually encounters weak comparability due to its independent training strategy, where different OCCs are trained with non-overlap training data from different tasks, leading to quite different distributions between their output. As shown in Figure 2, a sample of class "1" is mis-classified into class "2" due to the difference in output distributions of the two OOCs between class "1" and class "2". However, established POC-based methods focus more on the enhance discriminability rather than comparability.

To tackle the above problem, we propose \textbf{Dis}criminative and \textbf{C}omparable \textbf{O}ne-class classifiers for \textbf{I}ncremental \textbf{L}earning (DisCOIL).
The DisCOIL follows the POC framework but adopts variational auto-encoders (VAE \cite{vae}) as the one-class classifier instead of other well-established OCCs such as deep SVDD \cite{deepsvdd}, for the reason that VAE can act as a pseudo-sample generator in addition to being a one-class classifier.
With this advantage, DisCOIL trains each new-class VAE in contrast with old-class VAEs, which forces the new-class VAE to reconstruct better for new-class samples but worse for the old-class pseudo samples, thus enhancing the comparability. Furthermore, to ensure the discriminability, DisCOIL introduces a hinge reconstruction loss to improve the modeling robustness of each VAE for its corresponding class.
In brief, this work makes the following contributions:

\begin{itemize}
  \item We propose a novel parallel VAE architecture (DisCOIL) for incremental learning where each VAE acts as both a one-class classifier and a pseudo sample generator.
  \item With the generation ability of VAE, we introduce a classifier-contrastive loss and an inter-class loss to enhance the comparability between different OCCs.
  \item We propose a hinge reconstruction loss to improve the modeling robustness of each VAE for its corresponding class, which ensures the discriminability of DisCOIL.
  \item The proposed DisCOIL achieves state-of-the-art performance on several popular IL datasets.
\end{itemize}

The rest of this paper is organized as follows. 
Section~\ref{section2} introduces related works.
Section~\ref{section3} presents the proposed DisCOIL.
Section~\ref{section4} provides experimental comparison for DisCOIL with several state-of-the-art Class-IL methods.
Section~\ref{section-limit} discusses the limitaion of DisCOIL.
Finally, we summarize our work in Section~\ref{section5}.

\section{Related Work\label{section2}}

\subsection{Incremental learning}

Incremental learning (IL) requires models to learn new knowledge incrementally without forgetting \cite{survey1}.
According to the techniques used for preventing forgetting, IL methods are usually categorized into three categories: 
(1) regularization-based: methods use regularization terms to alleviate forgetting \cite{lwf,podnet, ewc, si, mas, adam-nscl};
(2) rehearsal-based: methods keep a fixed-size buffer to store samples of old tasks \cite{bic, gem, a-gem,der};
(3) dynamic architecture: methods utilize different network parameters for different tasks \cite{packnet, piggyback, pnn, cpg}.


Exemplar-free Class-IL is a setting of IL which adds constraints of inferencing without task identifications and training without the old-task-sample buffer.
Thus, exemplar-free Class-IL is extremely challenging because models need to learn to infer the task identification of a test sample implicitly or explicitly without any old-task data, and only a few methods can satisfy this setting.
This setting is significant because old-task data may not be available in some scenarios with data security or privacy restrictions.
To this end, some methods \cite{PGMA,hnet} generate network parameters according to the query sample.
Other methods realize multitasking classification by fusing the distribution of network parameters of different tasks \cite{IMM}.
Besides, OWM \cite{OWM} achieves exemplar-free Class-IL through orthogonal data projection innovatively and shows impressive performance.
Recently, some methods based on the framework of parallel one-class classifiers (POC) arise.
They train an OCC for each class which predicts the probability of a test sample belonging to the corresponding class.
ILCOC \cite{ilcoc} empolies a deep SVDD \cite{deepsvdd} model for each class.
And, per class learning (PCL) \cite{pcl} applies a sigmoid classifier as the basic OCC model.
Because of the special network structure, their parameter usage increases linearly with the number of learned classes.
However, the cost is worth it, both ILCOC and PCL show promising performance under exemplar-free Class-IL.
Compared with them, DisCOIL focuses more on the comparability among OCCs and shows better performance.

\subsection{Auto-encoder}

Auto-encoder \cite{ae} is a classic network structure that consists of an encoder and a decoder,
while the encoder encodes samples into latent vectors from the feature space to a latent space and the decoder reconstructs the latent vectors into reconstruction samples.
To deal with the dimensionality reduction task, during training, auto-encoders minimize the reconstruction error and learn to encode information as much as possible.
Since auto-encoder can only reconstruct in-distribution samples and can not reconstruct out-of-distribution samples,
auto-encoders are widely applied in anomaly detection \cite{arnet, memae}, with the reconstruction errors regarding as anomaly scores.
Besides, auto-encoders can generate samples by sampling latent vectors from latent space and decoding them through the decoder.
However, the latent space of the classic auto-encoder lacks regularity, which limits its generation ability.
Therefore, variational auto-encoder (VAE) \cite{vae} have been proposed and it encodes samples into a distribution. 
In addition, VAE introduces a regularization term that is a KL-divergence between that distribution and standard Gaussian distribution.
As a result, the latent space of VAE is more regular, and VAE has excellent generation ability.

\section{Method\label{section3}}

\subsection{Problem Setting and Method Overview\label{section31}}
Under the exemplar-free Class-IL setting, models need to learn a sequence of disjoint classification tasks incrementally.
Formally, let $Y^t$ represents the label set of task $t$, when $i \neq j$, $Y^i \cap Y^j = \emptyset$.
Assume the dataset of task $t$ is $D^t = \{(\mathbf{x}_k^t, y_k^t)\}^{n_t}_{k=1}$, where $n_t$ is the number of samples in $D^t$;
$\mathbf{x}_k^t$ and $y_k^t \in Y^t$ denote the $k$-th sample and the corresponding label, respectively.
During training, the model can only learn with the data from $D^t$;
during testing, we evaluate the model with the data from datasets of all learned tasks $\bigcup_{t=1}^{T}D^t$, where $T$ is the number of tasks.

\begin{figure}[t]
    \centering
    \includegraphics[width=0.9\columnwidth]{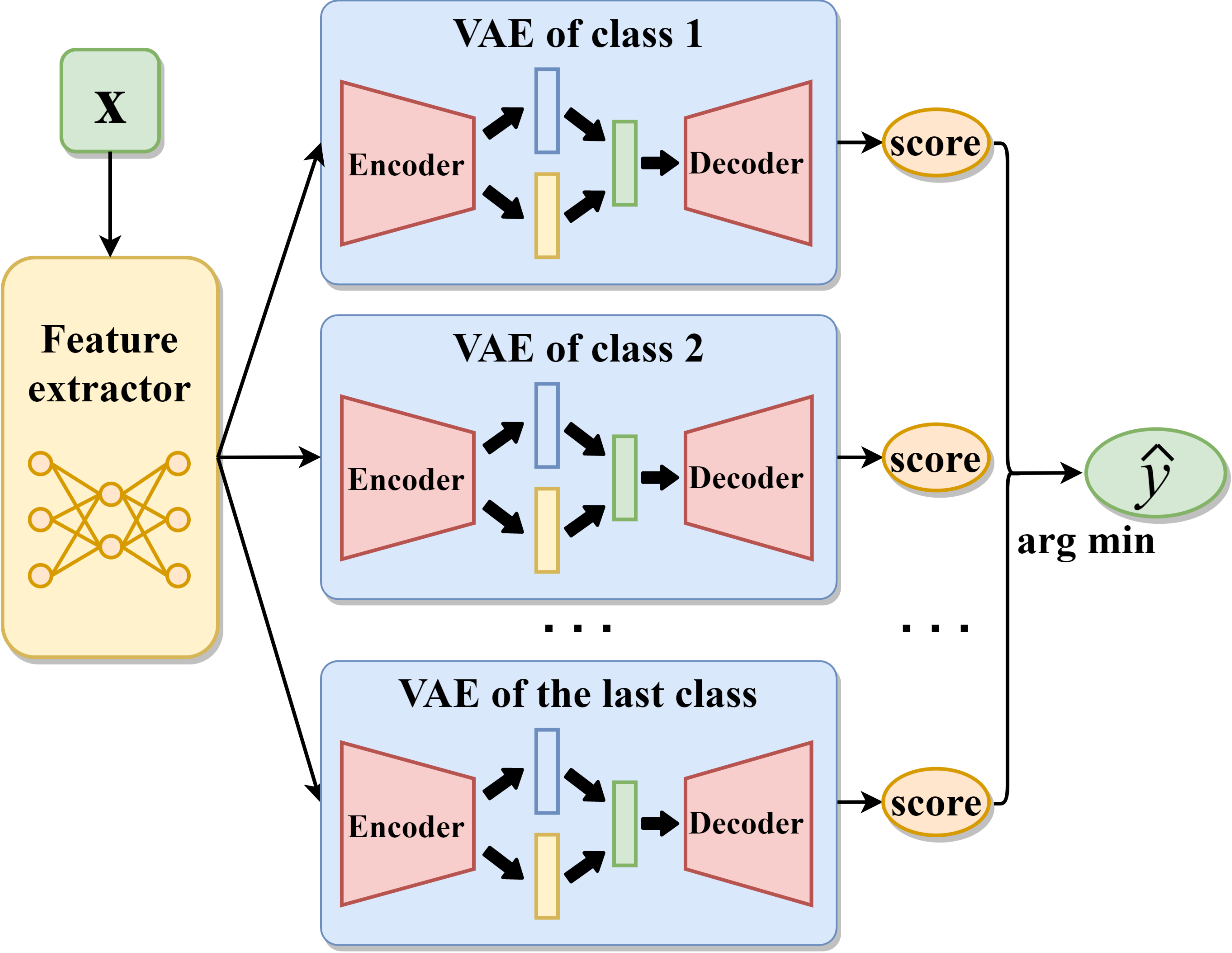} 
    \caption{
        The architecture of DisCOIL based on variational auto-encoder (VAE). 
    }
    \label{fig:architecture}
\end{figure}

\begin{figure*}[t]
  \centering
  \includegraphics[width=1.0\textwidth]{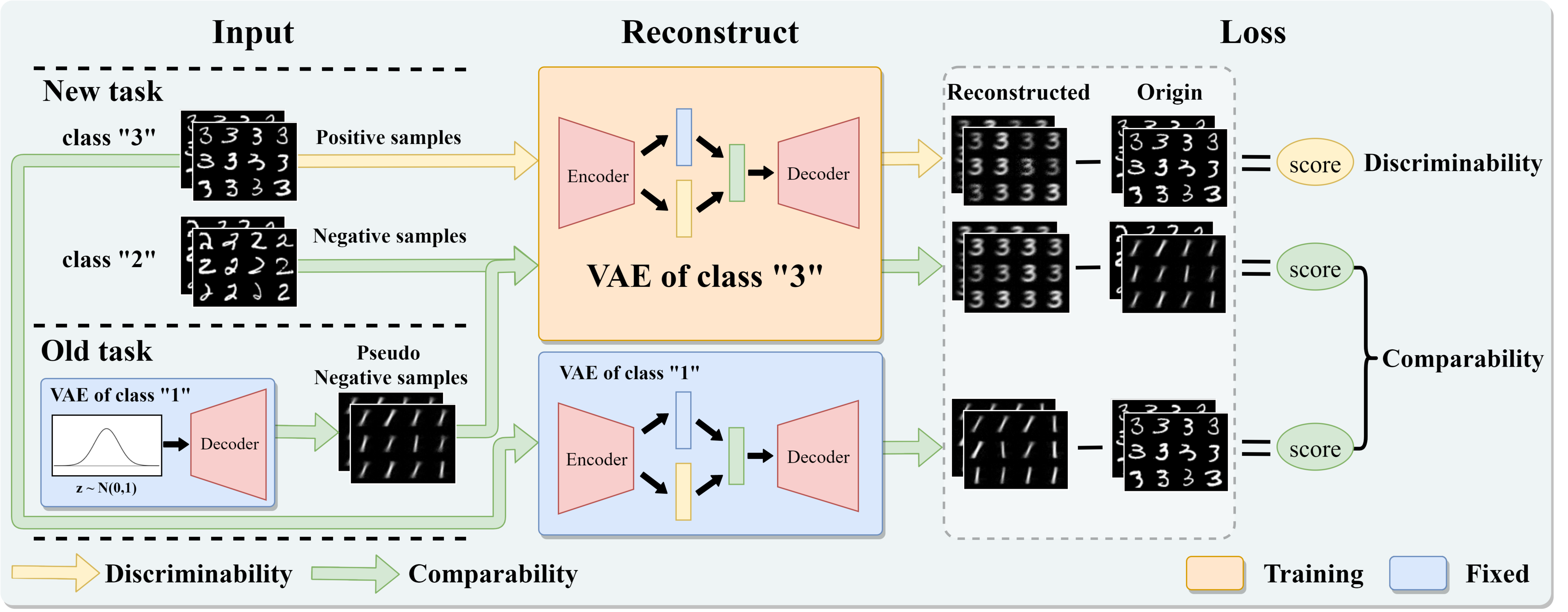} 
  \caption{
      Diagram of the training process in DisCOIL. 
      Assuming we are training the VAE corresponding to class ``3'', and the VAEs of class ``1'' and class ``2'' are trained before.
      For positive samples, we force the reconstruction errors to be lower than a pre-defined threshold $r_{intra}$;
      for negative samples, we require the reconstruction error to be greater than a pre-defined threshold $r_{inter}$.
      Besides, we also send the positive samples into old-task VAEs and force the new-class VAE to output lower anomaly scores than the old-task VAEs.
  }
  \label{fig:train}
\end{figure*}

Figure~\ref{fig:architecture} shows the overall architecture of DisCOIL, which consists of a pre-trained feature extractor and several VAE models.
Each VAE serves as both an OOC and a pseudo-sample generator for a specific class.
As an OCC, the reconstruction error reflects the degree of a given sample does not belonging to the corresponding classes, which thus is utilized as the anomaly score for one-class learning.
As a generator, VAE can generate pseudo samples which are employed to enhance the comparability among VAEs.
In addition, DisCOIL dynamically expands network capacity as the number of learned tasks increases.
When a new task $t$ comes, DisCOIL builds a new VAE model for each category in $Y^t$ and trains them one by one. 

We denote the VAE corresponding to class ${i}$ as $D_i(E_i(.))$,
where $D_i(.)$ and $E_i(.)$ are the encoder and the decoder, respectively.
For a query sample $\mathbf{x}$, ignoring the feature extraction process for simplicity, the anomaly score for class $i$ is calculated as
\begin{equation}
    score_i(\mathbf{x})=\|D_i(E_i(\mathbf{x})) - \mathbf{x}\|^2 ,\label{equation:score}
\end{equation}
and the final prediction result is determined by the class with the lowest anomaly score
\begin{equation}
    \hat{y} = \mathop{\arg\min}_{i} \{ score_i(\mathbf{x}) \} .\label{equation:result}
\end{equation} 

Since the anomaly score indicates the degree to which a given sample does not fall into this category corresponds to the VAE, the discriminability within each VAE is important for determining the intrinsic recognition ability of the model.
Moreover, the final prediction (Eq. (\ref{equation:result})) needs us to compare the anomaly scores of different VAEs, thus the comparability among VAEs is also essential to the performance of DisCOIL.
Furthermore, the prediction of an ideal model should satisfy the requirements shown in Figure~\ref{fig:comparability}. 

Taking the above considerations into account, we design the training process shown in Figure~\ref{fig:train}.
Assuming that we are training the VAE of class ``3", we regard the samples belonging to class ``3" as positive samples and the others as negative samples.
From the perspective of one-class learning, the discriminability of the VAE is enhanced by minimizing anomaly scores for positive samples.
Besides, we enhance the comparability among VAEs by training each new-class VAE in contrast with previously trained VAEs.
Specifically, the new-class VAE is required to reconstruct better for positive samples but worse for pseudo negative samples generated by the old-class VAEs.
Next, we elaborate on the processes of enhancing the discriminability within each VAE and the comparability among VAEs in Section~\ref{section32} and Section~\ref{section33}, respectively.

\subsection{Discriminability Enhancement within VAE\label{section32}}
As mentioned in Section~\ref{section1}, the performance of a POC-based model is affected by the discriminability within each OCC.
Namely, a discriminative OCC is expected to output low anomaly scores for positive samples while high anomaly scores for negative samples.
Thus, DisCOIL ensures the discriminability of VAEs by minimizing reconstruction errors for positive samples.

Classic reconstruction loss minimizes the mean squared error for all positive samples as small as possible, rendering the VAE susceptible to over-fitting (e.g. remembering all the training samples). Instead, under the IL scenario, the minimization of the reconstruction loss is sufficient as long as it can distinguish the positive samples from the negative samples.
Therefore, we propose intra-class loss to just enforce the reconstruction errors of positive samples to be smaller than an upper bound $r_{intra}$.
To be specific, when training the VAE of class $i$, the intra-class loss is defined as 
\begin{equation}
    L_{intra} = \frac{1}{|D^t_{i+}|} \sum_{\mathbf{x} \in D^t_{i+}} \max\{0, score_i(\mathbf{x})-r_{intra}\} \label{equation:intra},
\end{equation}
where $D^t_{i+} = \{\mathbf{x}_k^t|(\mathbf{x}_k^t, y_k^t) \in D^t, y_k^t = i\}$ denotes the positive sample set for class $i$, and $r_{intra} \geq 0$ is a hyperparameter shared by all VAEs.
The intra-class loss relaxes the requirements for easy-to-identify samples while pays more attention to hard-to-identify samples, which enhances the generalization ability as well as the discriminability of DisCOIL.

\subsection{Comparability Enhancement among VAEs\label{section33}}
As existing POC-based methods train OCCs with non-overlap training data from different tasks, the scores output by different OCCs may have quite different distributions, causing the weak comparability problem. 
To tackle this problem, we introduce the classifier-contrastive loss and the inter-class loss to train a new-class VAE in contrast with old-class VAEs.

For positive samples, the new-class VAE should reconstruct better than old-class VAEs.
To this end, we first acquire the reconstruction errors of the positive samples under old-class VAEs and then apply the classifier-contrastive loss to force the new-class VAE to output lower reconstruction errors than old-class VAEs, which can be formulated as 
\begin{equation}
    L_{CC} = \frac{1}{|D^t_{i+}|} \sum_{\mathbf{x} \in D^t_{i+}} \frac{1}{i} \sum_{j < i} \max\{0, score_i(\mathbf{x}) - score_j(\mathbf{x})\} ,
\end{equation}
where $score_j(\mathbf{x})$ represents the anomaly score output by the VAE corresponding to the old class $j$ where $j < i$.

For a negative sample, the new-class OCC should output a higher anomaly score than the corresponding old-class OCC.
Specifically, we propose inter-class loss to compel the reconstruction errors output by the new-class VAE for all negative samples higher than a lower bound $r_{inter}$:
\begin{equation}
    L_{inter} = \frac{1}{|D^t_{i-}|} \sum_{\mathbf{x}' \in D^t_{i-}} \max\{0, r_{inter} - score_i(\mathbf{x}')\}\label{equation:interloss} ,
\end{equation}
where $D^t_{i-}$ denotes negative sample set for class $i$, and $r_{inter}$ is a hyperparameter shared by all VAEs which satisfies $r_{inter} > r_{intra}$.

With regard to the negative sample set $D^t_{i-}$, note that we can only access the data of the current training task under the exemplar-free Class-IL scenario.
Nevertheless, the generation capability of VAE allows us to synthesize pseudo samples of old categories.
Specifically, random noises are sampled from the standard Gaussian distribution and fed into the decoders of an old-class VAE:
\begin{equation}
    \{D_j(z)| z \sim \mathcal{N}(0, 1)\} ,
\end{equation}
where $D_j(.)$ represents the decoder of the VAE corresponding to class $j$, and $z$ is a random noise.
Therefore the final negative sample set for class $i$ is defined as 
\begin{equation}
  \begin{aligned}
D^t_{i-} =& \{\mathbf{x}^t_k |(\mathbf{x}^t_k, y^t_k) \in D^t, y^t_k \neq i\} \cup  \\& \{D_j(z)| z \sim \mathcal{N}(0, 1), j \in Y^s, s < t\} .
  \end{aligned}
\end{equation} 

For a sample $\mathbf{x}'$ belonging to an old class $j$ ($j < i$), recall that the intra-class loss (Eq. (\ref{equation:intra})) forces 
\begin{equation}
  r_{intra} > score_j(\mathbf{x}') \label{equation:inter1}.
\end{equation}
In addition, minimizing Eq. (\ref{equation:interloss}) encourages 
\begin{equation}
  score_i(\mathbf{x}') > r_{inter} \label{equation:inter2}.
\end{equation}
Combining Eqs. (\ref{equation:inter1}) and (\ref{equation:inter2}) with $r_{inter} > r_{intra}$ implies:
\begin{equation}
  score_i(\mathbf{x}') > score_j(\mathbf{x}') \label{equation:inter3}.
\end{equation} 
Equation (\ref{equation:inter3}) indicates that jointly optimizing the intra-loss and the inter-loss indeed enhance the comparability between different OCCs. Furthermore, $r_{inter}-r_{intra}$ can be regarded as the comparability 
margin.    

Note that although the proposed intra-class loss and inter-class loss seem to share a similar form with the triplet loss \cite{tripletloss}. 
However, they are fundamentally different in terms of mechanism and function.
The triplet loss encourages the features from the same category to be similar while pushing features of different classes far away, which only applies to a single model. In contrast, the proposed intra and inter loss compel the output of different VAEs distributing in the same range, and the score gap between the positive and negative samples by a margin $r_{inter}-r_{intra}$. In other words, they are designed to enhance a cross-model property, i.e. the compatibility. 

Finally, the overall loss function is
\begin{equation}
    L = L_{intra} + \lambda_1 L_{CC} + \lambda_2 L_{inter} + L_{KL},
\end{equation}
where $\lambda_1$ and $\lambda_2$ are weight paremeters, and $L_{KL}$ is the regularization term in VAE \cite{vae} to pull the latent distribution to match the the standard Gaussian distribution.

\section{Experiments\label{section4}}

We evaluate the performance of the proposed DisCOIL in this section. All experiments are conducted on a workstation running OS Ubuntu 16.04 with 18 Intel Xeon 2.60GHz CPUs, 256 GB memory, and 6 NVIDIA RTX3090 GPUs.

\begin{table*}[t]
    \centering
    \newcolumntype{I}{!{\vrule width 1.3pt}}
    \newcommand\shline{\noalign{\global\savewidth\arrayrulewidth
                                \global\arrayrulewidth 1.2pt}%
                        \hline
                        \noalign{\global\arrayrulewidth\savewidth}}
    \renewcommand{\arraystretch}{1.1}
    \begin{tabular}{p{1.8cm}|p{2.0cm}<{\centering}Ip{1.7cm}<{\centering}|p{1.7cm}<{\centering}|p{1.7cm}<{\centering}|p{1.7cm}<{\centering}|p{1.7cm}<{\centering}}
        \shline
        Approach                & \textbf{Venues}  & \textbf{MNIST}     & \multicolumn{2}{c|}{\textbf{CIFAR10}} & \multicolumn{2}{c}{\textbf{Tiny-ImageNet}}  \\ 
        \shline 
        PTF                     &                  & w/o                & w/                & w/o               & w/                & w/o                     \\ 
        \shline                         
        Joint                   &                  & 97.68              & \multicolumn{2}{c|}{92.20}            &  \multicolumn{2}{c}{59.99}                 \\ 
        \hline                          
        Fine-tune               &                  & 19.94              & 23.73             & 19.62             &  8.35             & 7.92                    \\ 
        \shline                         
        LWF \cite{lwf}          &  PAMI'17         & 21.53              & 21.43             & 19.61 $^{\ddag}$  &  9.14             & 8.46 $^{\ddag}$         \\ 
        SI \cite{si}            &  ICML'17         & 20.81              & 27.43             & 19.48 $^{\ddag}$  &  8.95             & 6.58 $^{\ddag}$         \\ 
        IMM  \cite{IMM}         &  NeurIPS'17      & 67.25 $^{\dag}$    & -                 & 32.36 $^{\dag}$   &  -                & -                       \\ 
        iCaRL* \cite{icarl}     &  CVPR'17         & 71.41              & 71.15             & 49.02 $^{\ddag}$  &  24.82            & 7.53 $^{\ddag}$         \\
        oEWC \cite{oewc}        &  ICML'18         & 20.36              & 28.27             & 19.49 $^{\ddag}$  &  8.69             & 7.58 $^{\ddag}$         \\ 
        FDR* \cite{fdr}         &  ICLR'19         & 81.25              & 59.62             & 30.91 $^{\ddag}$  &  13.43            & 8.70 $^{\ddag}$         \\ 
        PGMA \cite{PGMA}        &  ICLR'19         & 81.70 $^{\dag}$    & -                 & 40.47 $^{\dag}$   &  -                & -                       \\ 
        OWM  \cite{OWM}         &  Nature'19       & 96.30 $^{\dag}$    & 75.39             & \textbf{52.83} $^{\dag}$&  40.29      & -                       \\ 
        HAL* \cite{hal}         &  AAAI'21         & 80.98              & 59.29             & 32.36 $^{\ddag}$  &  14.79            & -                       \\ 
        EBM  \cite{EBM}         &  ICLRW'21        & 53.12 $^{\dag}$    & -                 & 38.84 $^{\dag}$   &  -                & -                       \\ 
        ILCOC \cite{ilcoc}      &  CVPRW'21        & 86.05              & 69.04             & 38.40 $^{\dag}$   &  41.42            & 16.97 $^{\dag}$         \\ 
        PCL \cite{pcl}          &  AAAI'21         & 95.75              & 68.79             & -                 &  39.19            & -                       \\ 
        \textbf{DisCOIL}        &  This work       & \textbf{96.69}     & \textbf{77.35}    & 44.54             & \textbf{50.80}    & \textbf{19.75}          \\ 
        \shline
    \end{tabular}
    
    \caption{
        Incremental learning results for standard classification benchmarks.
        We report the average accuracy (\%) over five runs with different random seeds, and the higher is the better. 
        The row ``PTF'' indicates that models with (w/) or without (w/o) a pre-trained feature extractor.
        The results with ``\dag'' are quoted from the original paper, and the results with ``\ddag'' are quoted from \cite{der}.
        ``-'' indicates experiments we were unable to run, because of unavailable source code or intractable training.
        Note that the methods with ``*'' keep a buffer with a maximum size of 200 during incremental learning which can not be fairly compared with the other exemplar-free methods.
    }
    
    \label{table:compare}
\end{table*}

\subsection{Experimental Setting\label{section41}}

\textbf{Datasets}.
We select three widely used image classification datasets as our benchmarks and divide each of them into several incremental batches.
The statistics of these datasets are shown in the following:
\begin{itemize}
    \item \textbf{MNIST} \cite{mnist}: 
    a widely used handwritten digital character dataset which contains 60000 images of size $28 \times 28$ from 10 classes.
    In our experiments, we divide all classes into five incremental batches equally.
    Namely, each batch contains samples of two classes.
    \item \textbf{CIFAR10} \cite{cifar10}:
    a dataset contains 60000 images of size $32 \times 32$ from 10 classes.
    Similarly, we divide it into five incremental batches equally.
    \item \textbf{Tiny-ImageNet} \cite{tinyimagenet}:
    the largest dataset in our experiment with 200 classes, and there are 600 images for each class.
    We split the 200 classes into ten incremental batches, and each batch contains samples of 20 classes.
\end{itemize}

\textbf{Evalutaion metric}.
Following the setting in \cite{gem, der}, after learning all tasks, we evaluate models with the data of all learned tasks and report the average accuracy over five runs with random seeds.

\textbf{Baselines}.
We select several classic and latest state-of-the-art Class-IL methods:
(1) \textbf{LWF} \cite{lwf} uses knowledge distillation technique to regularize activations of neural network;
(2) \textbf{SI} \cite{si} regards contributions of loss reduction as the importance of network parameters and punishes their changes;
(3) \textbf{IMM} \cite{IMM} matches the moment of the posterior distribution of network parameters trained from different tasks;
(4) \textbf{iCaRL} \cite{icarl} uses samples in the buffer to calculate class centers;
(5) \textbf{oEWC} \cite{oewc} uses the fisher matrix to calculate the importance weights of network parameters and restricts their changes;
(6) \textbf{FDR} \cite{fdr} utilizes samples in the buffer to create a function space and measures the importance of network parameters with the function space;
(7) \textbf{PGMA} \cite{PGMA} generates network parameters according to query samples;
(8) \textbf{OWM} \cite{OWM} avoids forgetting through orthogonal data projection;
(9) \textbf{HAL} \cite{hal} keeps anchors for old tasks to alleviate forgetting;
(10) \textbf{EBM} \cite{EBM} solves the bias problem by using energy model;
(11) \textbf{ILCOC} \cite{ilcoc} applies parallel deep SVDD model architecture to deal with the Class-IL problem;
(12) \textbf{PCL} \cite{pcl} is a POC-based method that creates a sigmoid classifier for each class.
Besides, we also provide two base methods:
(13) \textbf{Joint} trains a neural network with all task data jointly, and its performance can be regarded as the upper bound of Class-IL methods;
(14) \textbf{Fine-tune}, the lower bound of Class-IL methods that directly fine-tune neural network parameters in each task without using any incremental learning strategy. 

\textbf{Implementation detail}.
We select vgg11 as the shared feature extractor for all methods as it has simple network architecture and low memory usage.
When applying the pre-trained feature extractor, DisCOIL can use a very shallow network as the VAE for each class, to keep their memory footprint low. As for the VAE in our method, both the encoder and decode are composed of three fully connected layers followed with a ReLU activation function. Although the size of DisCOIL will linearly increase with the number of classes, the overall storage overhead is acceptable as the size of each VAE is actually small, e.g. 0.98M for MNIST and 9.96M for the other two benchmarks.

\textbf{Hyper-parameter selection}.
We train each model with 10, 50, and 100 epochs for MNIST, CIFAR10, and Tiny-ImageNet, respectively.
Other hyperparameters are determined through a validation set which is obtained by sampling 10 percent samples from the training set. 
For DisCOIL, we choose Adam \cite{adam} to optimize the model with the weight decay factor of 0.001 for CIFAR10 and 0.01 for others. Both $\lambda_1$ and $\lambda_2$ are tuned from the range of $(0, 0.1, 0.2, 0.5, 1, 2, 5, 10)$; $r_{intra}$ is tuned from 0 to 100 and $r_{inter}$ is tuned from 50 to 1000.
More details can be found in our supplementary code.

\subsection{Comparison Results\label{section42}}

In this section, we compare the performance of our DisCOIL method with all baselines.
Table~\ref{table:compare} reports the comparision results on three classification datasets.
The results show that all methods benefit from the feature extractor because it reduces the difficulty of training.
Besides, the fine-tune reports poor average accuracies ($\approx 1/T$) on all datasets, which proves the existence of the forgetting in Class-IL.
Also, some regularization-based methods (oEWC, SI, LWF) fail since they can not deal with the bias problem of the softmax classifier \cite{setting}. 
With the help of a fixed-size buffer, rehearsal-based methods (iCaRL, FDR, HAL) can mitigate the bias problem and perform a competitive performance on MNIST and CIFAR10.
However, due to the fixed buffer size, they also fail on Tiny-ImageNet that contains 200 classes (equal to the buffer size), since only one sample is kept for each class and can not effectively simulate the distribution of each class to prevent forgetting.
Besides, all POC-based methods (ILCOC, PCL, and DisCOIL) obtain a promising performance under the exemplar-free Class-IL setting.

\begin{table}
  \begin{center}
      \newcolumntype{I}{!{\vrule width 1.5pt}}
      \newcommand\shline{\noalign{\global\savewidth\arrayrulewidth
                              \global\arrayrulewidth 1.2pt}%
                      \hline
                      \noalign{\global\arrayrulewidth\savewidth}}
      \renewcommand{\arraystretch}{1}
      \begin{tabular}{cIc|c|c|c}
          \shline
          Task No.  & \textbf{Var 1}  & \textbf{Var 2}  & \textbf{Var 3}    & \textbf{DisCOIL} \\ 
          \shline
          No. 1 & \textbf{97.43}  & 95.99           & 97.40          &  97.32  \\ 
          No. 2 & \textbf{91.56}  & 89.29           & 91.54          &  91.47  \\ 
          No. 3 & 82.00           & 80.53           & 82.08          &  \textbf{82.32}  \\ 
          No. 4 & 78.40           & 77.15           & 78.44          &  \textbf{78.46}  \\ 
          No. 5 & 76.97           & 75.74           & 76.89          &  \textbf{77.35}  \\ 

          \shline
      \end{tabular}
  \end{center}
  \caption{
      Average accuracy (\%) of ablation experiments on CIFAR10 with the pre-trained feature extractor. 
  }

  \label{table:ablation}
\end{table}

Compared with other methods, DisCOIL shows remarkable performance on all benchmarks.
Specifically, the accuracy of DisCOIL is 0.39\% and 2.78\% higher than the second-best performance on MNIST and Tiny-ImageNet without the feature extractor.
We notice that iCaRL and OWM show better performance than DisCOIL on CIFAR10, the reasons of which are two folds:
one is that iCaRL stores old task samples during incremental learning while DisCOIL does not, which gives iCaRL extra information to prevent forgetting;
the other is that the VAEs of DisCOIL are unable to generate high-quality samples on CIFAR10 without the feature extractor, resulting in the failure of the mechanisms for improving comparability among OCCs in DisCOIL.
With the help of the feature extractor, DisCOIL significantly outperforms other methods and achieves 4.37\% and 9.48\% higher performance than the second-best results on CIFAR10 and Tiny-ImageNet.
We highly recommend using DisCOIL equipping a pre-trained feature extractor, since it can not only reduce the difficulty of pseudo sample generation but also enable us to choose a smaller network for VAEs.

\subsection{Ablation Study\label{section43}}
In this section, we perform ablation studies to investigate the effectiveness of each component in DisCOIL on CIFAR10 with the feature extractor.
To this end, we construct several variations of DisCOIL to compare:
\textbf{Var 1}, which removes classifier-contrastive loss on the basis of DisCOIL by setting $\lambda_1 = 0$;
\textbf{Var 2}, which sets $\lambda_2 = 0$ to remove the inter-class loss;
\textbf{Var 3}, which is based on DisCOIL and does not use the pseudo samples generated by VAEs.
The results are shown in Table~\ref{table:ablation}, and we can see all parts of DisCOIL contribute to its final performance.

\textbf{Effectiveness of classifier-contrastive loss.}
We first analyze the effectiveness of the classifier-contrastive loss by comparing Var 1 and DisCOIL.
When learning task 1, the absence of the old-task OCC makes the classifier-contrastive loss useless, which may result in Var 1 outperforming DisCOIL;
As the number of learned tasks increases, DisCOIL gradually outperforms Var 1.
This phenomenon indicates that the classifier-contrastive loss can only improve the comparability among VAEs trained from different tasks, since the VAEs from the same task already have enough comparability with the help of inter-class loss.

\textbf{Effectiveness of inter-class loss.}
As illustrated in Table~\ref{table:ablation}, the performance of DisCOIL is greater than Var 2 at any incremental stage.
The promotion comes from the following two aspects: (1) by utilizing negative samples from each task dataset, the inter-class loss can improve the comparability among VAEs from the same task;
(2) the result of Var 3 shows that the pseudo samples generated by old-task VAEs can assist the inter-class loss to improve the comparability among VAEs from different tasks.

\begin{figure}[t]
  \centering
  \includegraphics[width=0.98\columnwidth]{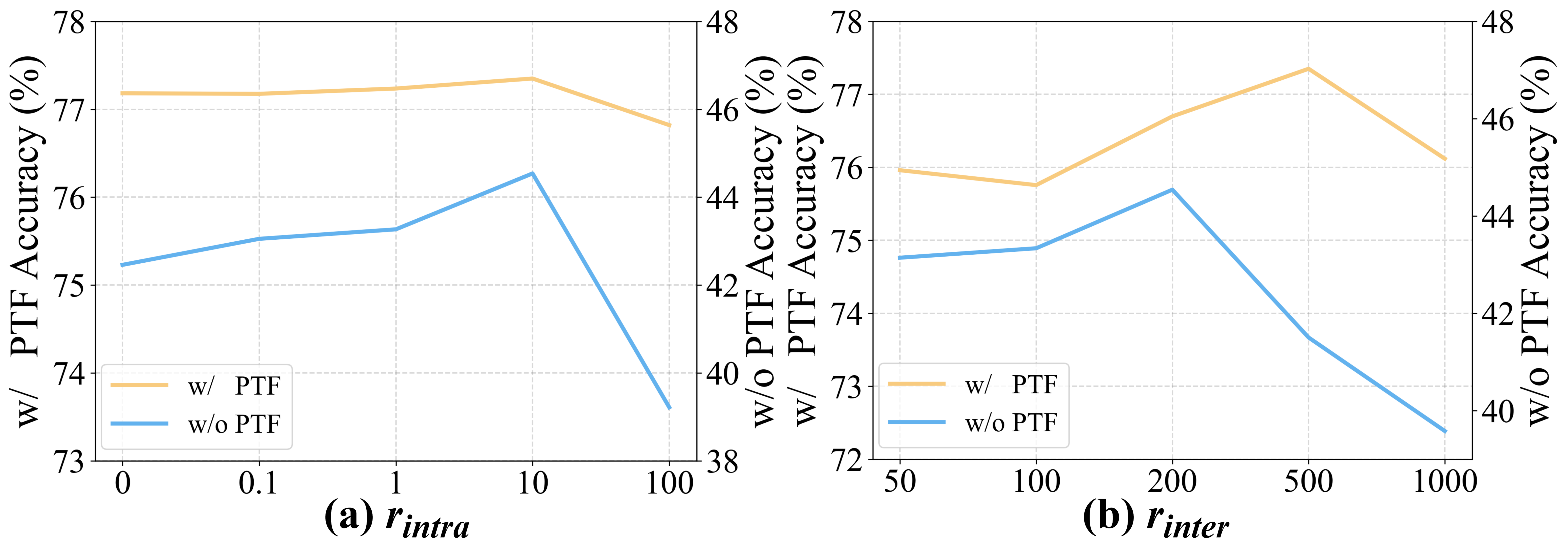} 
  \caption{
    Sensitivity analysis of $r_{intra}$ (a) and $r_{inter}$ (b) on CIFAR10. 
  }
  \label{fig:sensitive}
\end{figure}

\subsection{Sensitivity Analysis\label{section44}}

\textbf{Sensitivity analysis of $r_{intra}$.}
In this section, we investigate the influence of the upper bound $r_{intra}$ in the intra-class loss.
We select five different values, including 0, 0.1, 1, 10, 100.
As illustrated in Figure~\ref{fig:sensitive} (a), the model has a poor performance when $r_{intra}=0$ since the intra-class loss is degraded into MSE loss. 
Besides, as $r_{intra}$ increasing, the average accuracy grows and reaches the peak when $r_{intra}=10$.
When $r_{intra}=100$, the performance drops again because VAEs do not learn to decrease the reconstruction error and lack of discriminability for one-class classification.
In addition, we notice that $r_{intra}$ has little effect on model performance when applying the pre-trained feature extractor because the feature extractor can smooth the input of VAEs and alleviate over-fitting.

\textbf{Sensitivity analysis of $r_{inter}$.}
In addition, we evaluate the effect of the lower bound in inter-class loss $r_{inter}$.
We also consider five different values that greater than $r_{intra}$, including 50, 100, 200, 500, 1000.
As illustrated in Figure~\ref{fig:sensitive} (b), with $r_{inter}$ increasing, the average accuracy is also gradually growing.
Because more samples satisfy Eq. (\ref{equation:inter1}-\ref{equation:inter3}) and ensure high comparability between different OCCs when the margin $r_{inter} - r_{intra}$ becomes larger.
It can observed the performance drops when $r_{inter}=1000$, since a large $r_{inter}$ makes the model occupies excessive network capacity to reconstruction worse on old-class samples.
Besides, when applying the feature extractor, the optimal value of $r_{inter}$ is greater than that without the feature extractor, the reason is that the feature extractor can reduce the difficulty of reconstructing negative samples and meanwhile does not hurt the performance of VAEs.
 
\begin{figure}[t]
  \centering
  \includegraphics[width=0.8\columnwidth]{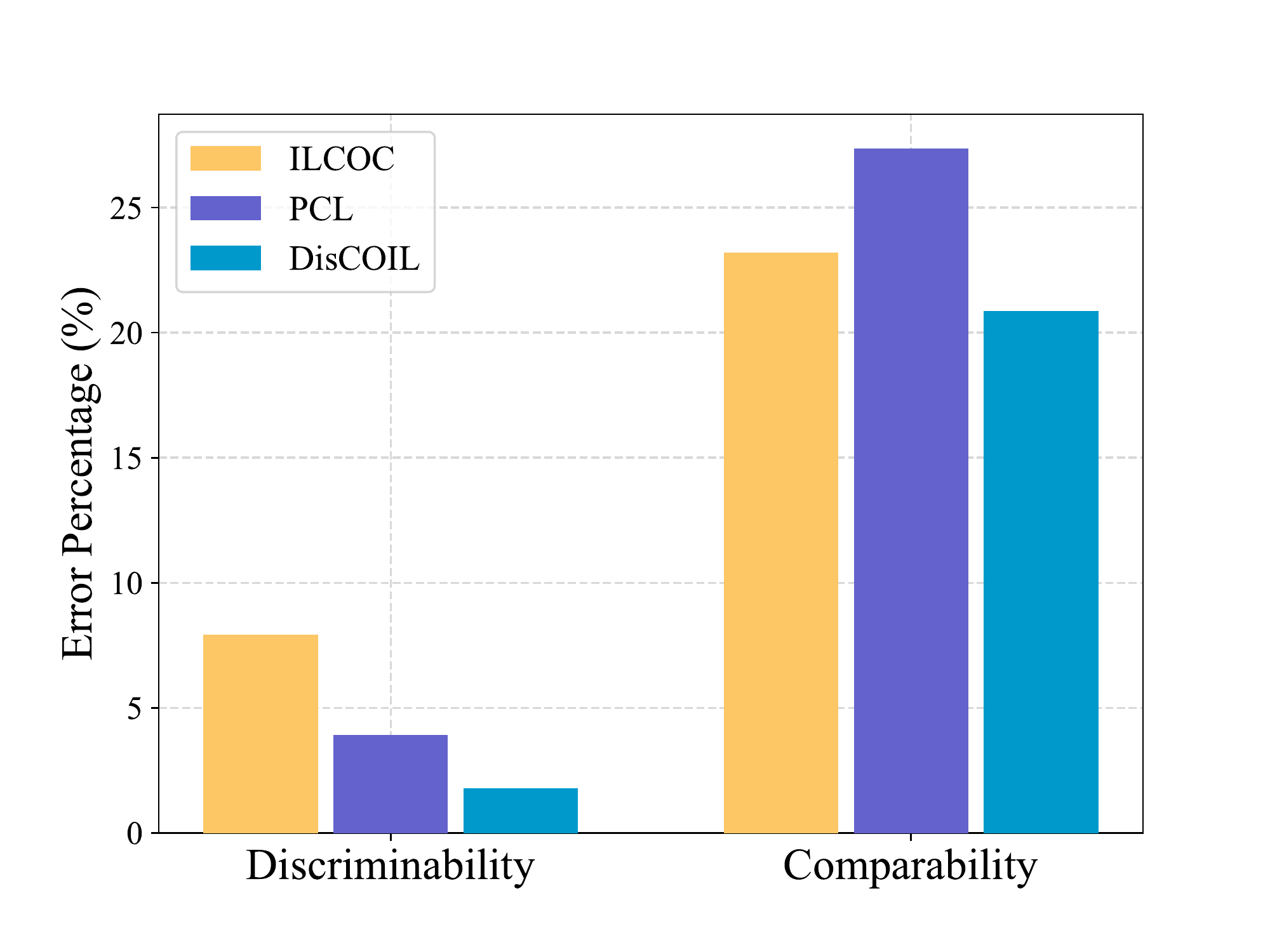} 
  \caption{
      Statistical results of the misclassifications on CIFAR10 with the pre-trained feature extractor. 
      All misclassified samples are divided into two categories, namely, caused by lacking comparability or discriminability.
      We report the proportion of misclassified samples to the total test samples, and the lesser is the better.
  }
  \label{fig:compareresult}
\end{figure}

\subsection{Misclassification Analysis \label{section45}}

As mentioned in Section~\ref{section1}, the performance of POC-based methods depends on the discriminability and the comparability of OCCs.
This section analyzes the misclassifications of DisCOIL, ILCOC, and PCL.
Specifically, for a misclassified sample $\mathbf{x}$ and its label $y$, we first feed the scores of all OCCs into the softmax function to convert them into probabilities.
If the OCC corresponding to class $y$ predicts a too low probability (lesser than $1/(\sum_t{|Y^t|})$ in our experiment), we believe that the misclassification is caused by the low discriminability. 
Otherwise, the low comparability among OCCs leads to this misclassification.
Figure~\ref{fig:compareresult} reports the statistical results of all misclassified samples, and we can see that all POC-based methods suffer from lacking comparability among OCCs.
Besides, the OCCs of DisCOIL have more discriminability and comparability than others.

\section{Discussion\label{section-limit}}
DisCOIL enjoys the advantages of the POC framework, but it also inherits partial flaws of POC.
One drawback is that the parameter usage of DisCOIL linearly increases with the number of learned classes.
To reduce the impact of excessive network parameters, we recommend equipping DisCOIL with a pre-trained feature extractor, since it can not only improve the model performance (see Table~\ref{table:compare}) but also allow a smaller network to implement VAEs.
Another feasible approach is to combine DisCOIL with the mask weights learning techniques in Task-IL methods, such as Piggyback \cite{piggyback}, CPG \cite{cpg} and PackNet \cite{packnet}, by taking each VAE training as an independent task. These techniques enable DisCOIL to adapt a single network to multiple one-class classifiers.
As this paper focuses on solving the weak discriminability and comparability of POC, we leave the above exploration for future work. 

\section{Conclusion\label{section5}}
In this paper, to solve the weak discriminability and comparability of the POC IL framework, we propose a novel IL method DisCOIL based on parallel VAE architecture. With the advantage of VAE, a joint loss is designed to compel DisCOIL to satisfy a cross-model relation, which enhances the comparability between different OCCs. 
Furthermore, DisCOIL introduces a hinge reconstruction loss to improve the discriminability.
We evaluate DisCOIL on several classification datasets, including MNIST, CIFAR10, and Tiny-ImageNet. The experimental results show that DisCOIL achieves state-of-the-art performance.

Some further studies are left in the future.
Firstly, we will try to integrate the mask weights learning techniques of Task-IL methods into DisCOIL to reduce the parameter usage.
Secondly, we will investigate the distribution alignment technique to improve the comparability among OCCs and the performance of POC-based methods.

{\small
\bibliographystyle{ieee_fullname}
\bibliography{egbib}
}

\end{document}